\def\year{2020}\relax
\documentclass[letterpaper]{article} % DO NOT CHANGE THIS
\usepackage{aaai20}  % DO NOT CHANGE THIS
\usepackage{times}  % DO NOT CHANGE THIS
\usepackage{helvet} % DO NOT CHANGE THIS
\usepackage{courier}  % DO NOT CHANGE THIS
\usepackage[hyphens]{url}  % DO NOT CHANGE THIS
\usepackage{graphicx} % DO NOT CHANGE THIS
\usepackage{graphicx}  % DO NOT CHANGE THIS
\usepackage{amsmath}
\usepackage{multirow}
\usepackage{bm}
\usepackage{amssymb}
\usepackage{threeparttable}

\usepackage[T1]{fontenc}
\usepackage{textcomp}

\title{Cross-Modality Paired-Images Generation for RGB-Infrared \\ Person Re-Identification}
\author{Guan-An Wang,\textsuperscript{\rm 1}\textsuperscript{\rm 2}
Tianzhu Zhang,\textsuperscript{\rm 4}
Yang Yang,\textsuperscript{\rm 1}
Jian Cheng,\textsuperscript{\rm 1}\textsuperscript{\rm 2}\textsuperscript{\rm 3} \\
\Large \textbf{Jianlong Chang,\textsuperscript{\rm 1}\textsuperscript{\rm 2}
Xu Liang,\textsuperscript{\rm 1}\textsuperscript{\rm 2}
and Zengguang Hou\textsuperscript{\rm 1}\textsuperscript{\rm 2}\textsuperscript{\rm 3}\thanks{corresponding author}
} \\
\textsuperscript{\rm 1} Institute of Automation, Chinese Academy of Sciences, Beijing, China \\
\textsuperscript{\rm 2} University of Chinese Academy of Sciences, Beijing, China \\
\textsuperscript{\rm 3} Center for Excellence in Brain Science and Intelligence Technology, Beijing, China \\
\textsuperscript{\rm 4} University of Science and Technology of China, Beijing, China\\
\{wangguanan2015,liangxu2013,zengguang.hou\}@ia.ac.cn, tzzhang@ustc.edu.cn, \\ \{yang.yang,jcheng,jianlong.chang\}@nlpr.ia.ac.cn
}

\begin{document}

\maketitle

\begin{abstract}

RGB-Infrared (IR) person re-identification is very challenging due to the large cross-modality variations between RGB and IR images.
The key solution is to learn aligned features to the bridge RGB and IR modalities.
However, due to the lack of correspondence labels between every pair of RGB and IR images, most methods try to alleviate the variations with set-level alignment by reducing the distance between the entire RGB and IR sets. However, this set-level alignment may lead to misalignment of some instances, which limits the performance for RGB-IR Re-ID.
%
%Considering no correspondence labels between every pair of RGB and IR images, most methods try to alleviate the variations with set-level alignment by reducing marginal distribution divergence between the entire RGB and IR sets. However, this set-level alignment strategy may lead to misalignment of some instances, which limit the performance for RGB-IR Re-ID.
%
Different from existing methods, in this paper, we propose to generate cross-modality paired-images and perform both global set-level and fine-grained instance-level alignments.
Our proposed method enjoys several merits.
First, our method can perform set-level alignment by disentangling modality-specific and modality-invariant features. Compared with conventional methods, ours can explicitly remove the modality-specific features and the modality variation can be better reduced.
Second, given cross-modality unpaired-images of a person, our method can generate cross-modality paired images from exchanged images. With them, we can directly perform instance-level alignment by minimizing distances of every pair of images.
Extensive experimental results on two standard benchmarks demonstrate that the proposed model favourably against state-of-the-art methods. Especially, on SYSU-MM01 dataset, our model can achieve a gain of $9.2\%$ and $7.7\%$ in terms of Rank-1 and mAP.
Code is available at \textit{https://github.com/wangguanan/JSIA-ReID}.

\end{abstract}

\section{Introduction}

\begin{figure}[t]
\center
\includegraphics[width=\linewidth]{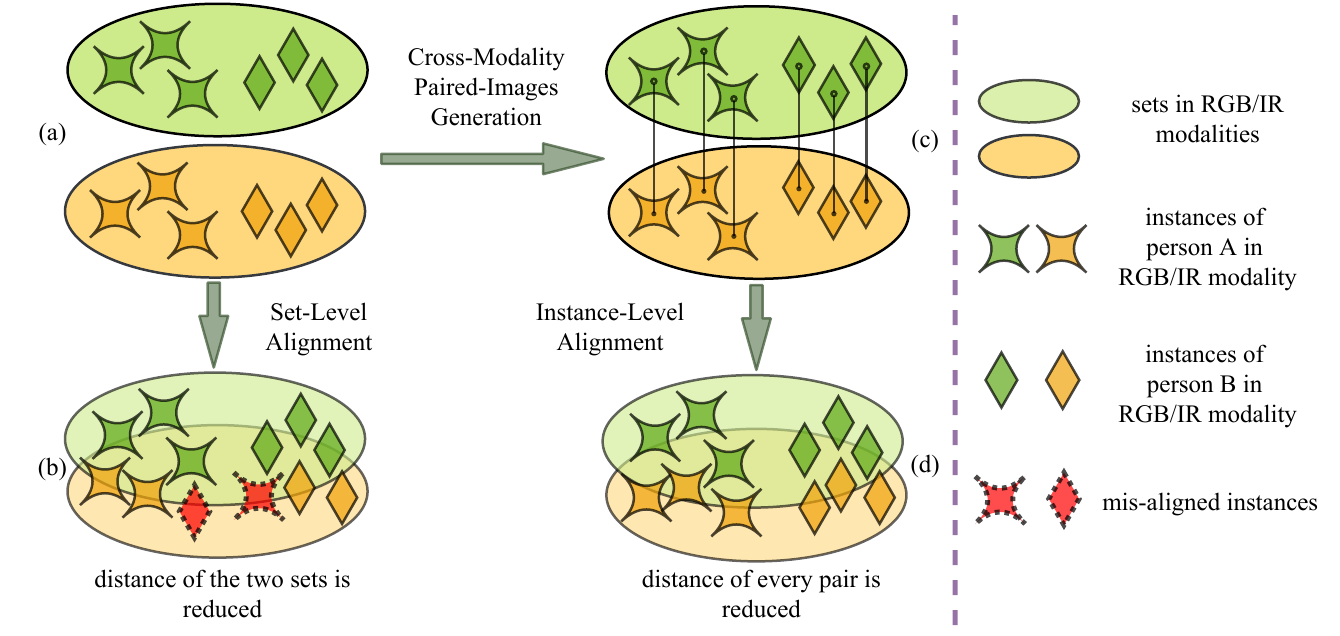}
\caption{
Illustration of set-level and instance-level alignment (please view in color). (a) There is a significant gap between the RGB and IR sets. (b) Existing methods perform set-level alignment by minimizing distances between the two sets, which may lead to misalignment of some instances. (c) Our method first generates cross-modality paired-images. (d) Then, instance-level alignment is performed by minimizing distances between each pair of images.
}
\label{fig:set_vs_instance}
\end{figure}

\begin{figure}[t]
\center
\includegraphics[width=\linewidth]{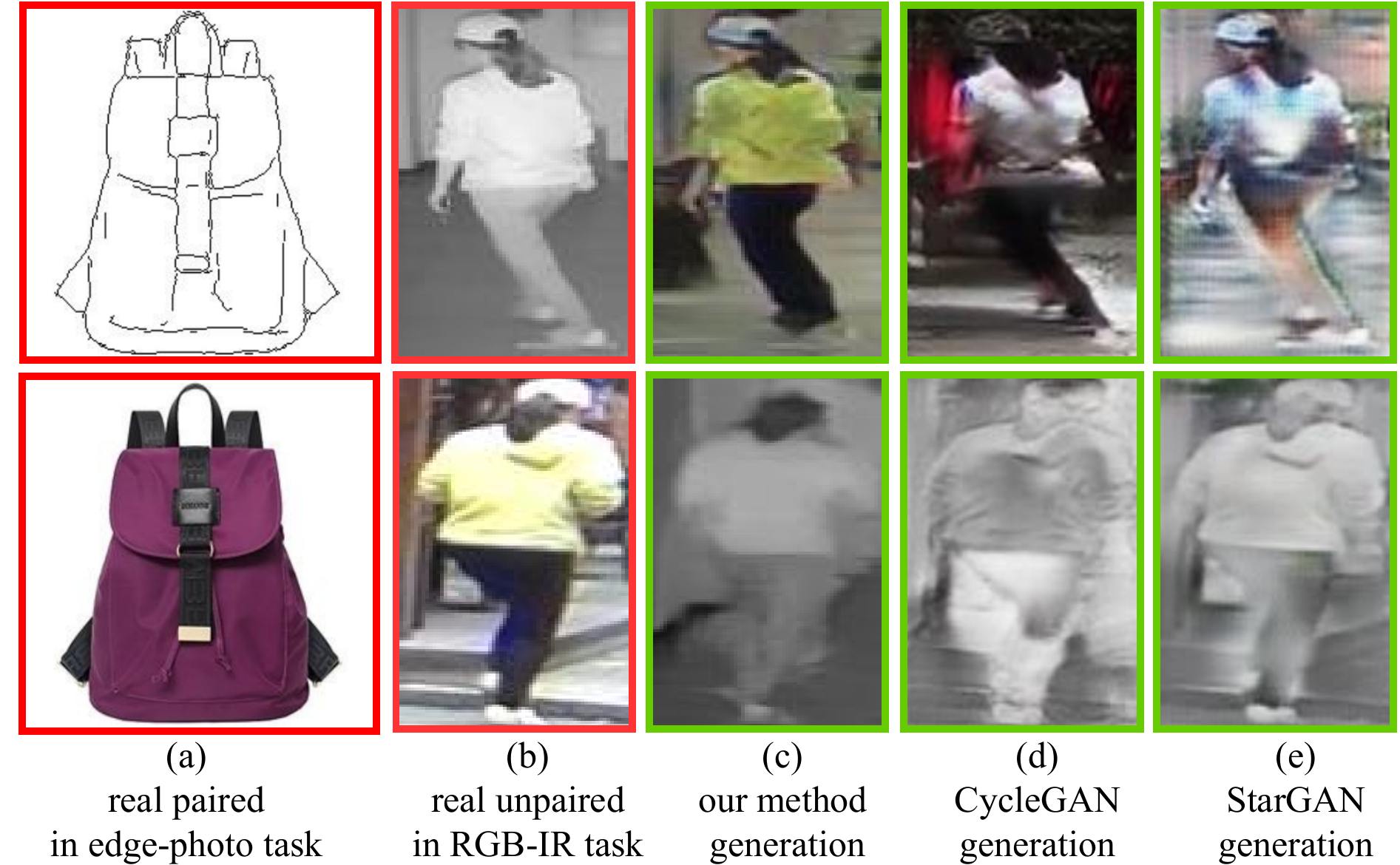}
\caption{(a) In the edge-photo task, we can get cross-modality paired-images. By minimizing their distances in a feature space, we can easily reduce the cross-modality gap. (b) In RGB-IR Re-ID task, we have only unpaired-images. The appearance variation caused by the cross-modality gap makes the task more challenging. (c) Our method can well generate images paired with given ones, which help us to improve RGB-IR Re-ID. (d,e) Vanilla image translation models such as CycleGAN \cite{zhu2017unpaired} and StarGAN \cite{choi2018stargan} fail to deal with this issue.
}
\label{fig:images_comparision}
\end{figure}

Person Re-Identification (Re-ID) \cite{gong2014person,zheng2016person} is widely used in various applications such as video surveillance, security and smart city. Given a query image of a person, Re-ID aims to find images of the person across disjoint cameras. It's very challenging due to the large intra-class and small inter-class variations caused by different poses, illuminations, views, and occlusions.
%
%To tackle the above issue, lots of methods have been proposed, which can be grouped into hand-crafted descriptors \cite{ma2014covariance,yang2014salient,liao2015person}, metric learning \cite{zheng2013reidentification,koestinger2012large,liao2015efficient}, and deep learning \cite{zheng2016person,hermans2017defense,sun2018beyond}. 
%
Most of existing Re-ID methods focus on visible cameras and RGB images, and formulate the person Re-ID as a single-modality (RGB-RGB) matching problem.

However, the visible cameras are difficult in capturing valid appearance information under poor illumination environments (e.g. at night), which limits the applicability of person Re-ID in practical. Fortunately, most surveillance cameras can automatically switch from visible (RGB) to near-infrared (IR) mode, which facilitates such cameras to work at night.
Thus, it is necessary to study the RGB-IR Re-ID in real-world scenarios, which is a cross-modality matching problem.
Compared with RGB-RGB single-modality matching, RGB-IR cross-modality matching is more difficult due to the large variation between the two modalities. As shown in  Figure \ref{fig:images_comparision}(b), RGB and IR images are intrinsically distinct and heterogeneous, and have different wavelength ranges. 
Here, RGB images have three channels containing color information of visible light, while IR images have one channel containing information of invisible light. 
%As a result, even human can hardly recognize the person cross the two modalities.

The key solution is to learn aligned features to bridge the two modalities. However, due to the lack of correspondence labels between every pair of images in different modalities like in Figure \ref{fig:images_comparision}(a), 
existing RGB-IR Re-ID methods \cite{wu2017rgb,ye2018hierarchical,ye2018visible,dai2018cross,hao2019hsme} try to reduce the marginal distribution divergence between RGB and IR modalities, while cannot deal with their joint distributions.
That is to say, as shown in Figure \ref{fig:set_vs_instance}(b), they only focus on the global set-level alignment between the entire RGB and IR sets while neglecting the fine-grained instance-level alignment between every two images.
This may lead to misalignment of some instances when performing the global alignment \cite{chen2018re}.
Although we can alleviate this issue by using label information, in Re-ID task, labels of training and test sets are unshared. Thus, simply fitting training labels may not perform very well for unseen test labels.

%
\iffalse
In~\cite{wu2017rgb}, Wu \textit{et al.}  collect a cross-modality RGB-IR dataset named SYSU RGB-IR Re-ID. The proposed method explores three different network structures and uses deep zero-padding for training one-stream network towards automatically evolving domain-specific nodes in the network for cross-modality matching.
%
Ye \textit{et al.}~\cite{ye2018hierarchical} propose a hierarchical cross-modality matching model by jointly optimizing the modality-specific and modality-shared metrics. The modality-specific metrics transform two heterogenous modalities into a consistent space that modality-shared metric can be subsequently learnt.
In~\cite{ye2018visible}, a dual-path network is proposed  with a new bi-directional dual-constrained top-ranking loss to learn discriminative feature representations.
%
In~\cite{dai2018cross},  Dai \textit{et al.} introduce a  cross-modality generative adversarial network (cmGAN)  to reduce the distribution divergence of RGB and IR features.
%
Very recently, Hao \textit{et al.}~\cite{hao2019hsme} achieve visible thermal person re-identification via a hyper-sphere manifold embedding model.
\fi

%
Different from the existing approaches, a heuristic method is to use cross-modality paired-images in Figure \ref{fig:images_comparision}(a). With the paired images, we can directly reduce the instance-level gap by minimizing the distance between every pair of images in a feature space.
However, as in Figure \ref{fig:images_comparision}(b),all images are un-paired in RGB-IR Re-ID task. This is because the two kinds of images are captured at different times. RGB images are captured at daytime while IR ones at night.
We can also translate images from one modality to another by using image translation models, such as CycleGAN \cite{zhu2017unpaired} and StarGAN \cite{choi2018stargan}. But these image translation models can only learn one-to-one mappings, while mapping from IR to RGB images are one-to-many. For example, gray in IR mode can be blue, yellow even red in RGB mode.
Under this situation, CycleGAN and StarGAN often generate some noisy images and cannot be used in the following Re-ID task.
As shown in Figure \ref{fig:images_comparision}(d,e), the generated images by CycleGAN and StarGAN are unsatisfying.

To solve the above problems, in this paper, we propose a novel Joint Set-level and Instance-Level Alignment Re-ID (JSIA-ReID) which enjoys several merits.
First, our method can perform set-level alignment by disentangling modality-specific and modality-invariant features. Compared with encoding images with only one encoder, ours can explicitly remove the modality-specific features and significantly reduce the modality-gap.
Second, given cross-modality unpaired-images of a person, our method can generate cross-modality paired-images. With them, we can directly perform instance-level alignment by minimizing the distances between the two images in a feature space. The instance-level alignment can further reduce the modality-gap and avoid misalignment of instances.

\begin{figure*}[t]
\center
\includegraphics[width=\linewidth]{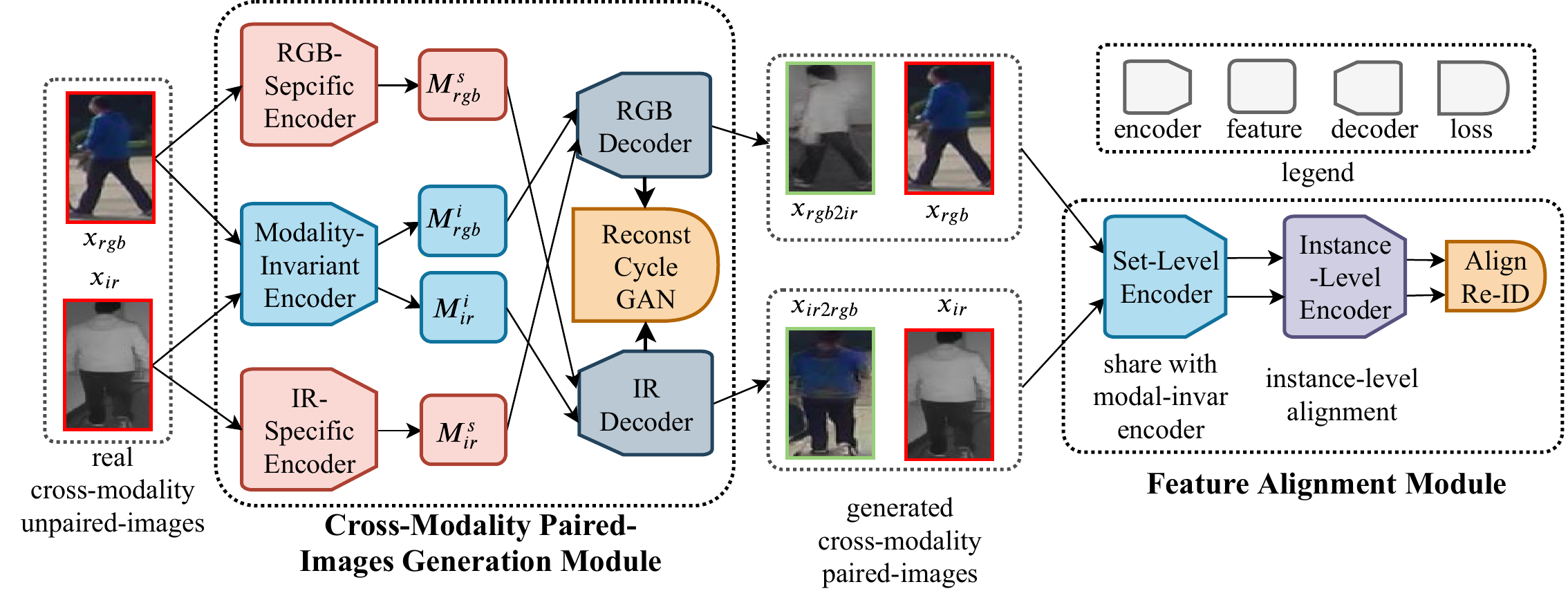}
\caption{Our proposed framework consists of a cross-modality paired-images generation module $\mathcal{G}$ and a feature alignment module $\mathcal{F}$. 
$\mathcal{G}$ first disentangle images to modality-specific and modality-invariant features, and then decode from the exchanged features. 
$\mathcal{F}$ first use the modality-invariant encoder to perform set-level alignment, then further perform instance-level alignment by minimizing distance of each pair images. 
Finally, by training the two modules with re-id loss, we can learn both modality-aligned and identity-discriminative features.
}
\label{fig:framework}
\end{figure*}

Specifically, as shown in Figure \ref{fig:framework}, our proposed method consists of a generation module $\mathcal{G}$ to generate cross-modality paired-images and a feature alignment module $\mathcal{F}$ to learn both set-level and instance-level aligned features.
The generation module $\mathcal{G}$ includes three encoders and two generators. The three encoders disentangle a RGB(IR) image to modality-invariant and RGB(IR) modalities-specific features. Then, the RGB(IR) decoder takes a modality-invariant feature from an IR(RGB) image and a modality-specific feature from an IR(RGB) image as input. By decoding from the across-feature, we can generate cross-modality paired-images as in Figure \ref{fig:images_comparision}(c).
In the feature alignment module $\mathcal{F}$, we first utilize an encoder whose weights are shared with modality-invariant encoder. It can map images from different modalities into a shared feature space. Thus, set-level modality-gap can be significantly reduced. Then, we further import an encoder to refine the features to reduce the instance-level modality-gap by minimizing distance between feature maps of every pair of cross-modality images.
Finally, by jointly training the generation module $\mathcal{G}$ and feature alignment module $\mathcal{F}$ with the re-id loss, we can learn both modality-aligned and identity-discriminative features.

The major contributions of this work can be summarized as follows.
(1) We propose a novel method to generate cross-modality paired-images by disentangling features and decoding from exchanged features. To the best of our knowledge, it is the first work to generate cross-modality paired-images for the RGB-IR Re-ID task.
(2) Our method can simultaneously and effectively reduce both set-level and instance-level modality-variation.
(3) Extensive experimental results on two standard benchmarks demonstrate that the proposed model performs favourably against state-of-the-art methods.

\section{Related Works}

%In this section, we briefly overview methods that are related to RGB-RGB person re-identification,  RGB-IR person re-identification and generative adversarial networks.

\textbf{RGB-RGB Person Re-Identification.} RGB-RGB person re-identification addresses the problem of matching pedestrian RGB images across disjoint visible cameras \cite{gong2014person}. 
Recently, many deep ReID methods \cite{zheng2016person,hermans2017defense,wang2019color} have been proposed.
Zheng \textit{et al.} \cite{zheng2016person} learn identity-discriminative features by fine-tuning a pre-trained CNN to minimize a classification loss.  In \cite{hermans2017defense},  Hermans  \textit{et al.} show that using a variant of the triplet loss outperforms most other published methods by a large margin.
%In \cite{sun2018beyond}, a network named Part-based Convolutional Baseline (PCB) is proposed to learn fine-grained part-level features with a uniform partition strategy.
%
Most of exiting methods focus on the RGB-RGB Re-ID task, and cannot perform well for the RGB-IR Re-ID task, which limits the applicability in practical surveillance scenarios.

\textbf{RGB-IR Person Re-Identification.}
RGB-IR Person re-identification attempts to match RGB and IR images of a person under disjoint cameras. Besides the difficulties of RGB-RGB Re-ID, RGB-IR Re-ID faces a new challenge due to cross-modality variation between RGB and IR images.
%
% In~\cite{wu2017rgb}, Wu \textit{et al.}  collect a cross-modality RGB-IR dataset named SYSU RGB-IR Re-ID. The proposed method explores three different network structures and uses deep zero-padding for training one-stream network towards automatically evolving domain-specific nodes in the network for cross-modality matching.
%
In~\cite{wu2017rgb}, Wu \textit{et al.}  collect a cross-modality RGB-IR dataset named SYSU RGB-IR Re-ID and explores three different network structures with zero-padding for automatically evolve domain-specific nodes in the network.
%
%Ye \textit{et al.}~\cite{ye2018hierarchical} propose a hierarchical cross-modality matching model by jointly optimizing the modality-specific and modality-shared metrics. The modality-specific metrics transform two heterogeneous modalities into a consistent space that modality-shared metric can be subsequently learnt.
%In~\cite{ye2018visible}, a dual-path network is proposed with a new bi-directional dual-constrained top-ranking loss to learn discriminative feature representations.
%
Ye \textit{et al.} utilize a dual-path network with a bi-directional dual-constrained top-ranking loss \cite{ye2018hierarchical} and modality-specific and modality-shared metrics \cite{ye2018visible}.
In~\cite{dai2018cross},  Dai \textit{et al.} introduce a  cross-modality generative adversarial network (cmGAN)  to reduce the distribution divergence of RGB and IR features.
Hao \textit{et al.}~\cite{hao2019hsme} achieve visible thermal person re-identification via a hyper-sphere manifold embedding model.
In \cite{Wang_2019_ICCV} and \cite{wang2019learning}, they reduce modality-gap in both image and feature domains.
Most above methods mainly focus on global set-level alignment between the entire RGB and IR sets, which may lead to misalignment of some instances.
Different from them, our proposed method performs both global set-level and fine-grained instance-level alignment, and achieves better performance.

\textbf{Person Re-Identification with GAN.}
Recently, many methods attempt to utilize GAN to generate training samples for improving Re-ID.
Zheng \textit{et al.} \cite{zheng2017unlabeled} use a GAN model to generate unlabeled images as data augmentation.
%
%Huang \textit{et al.} \cite{huang2018multi} first assign pseudo labels to generated pedestrian images and then learn them in a supervision manner.
%
Zhong \textit{et al.} %\cite{zhong2018camera,zhong2018generalizing,zhong2019invariance} 
\cite{zhong2018camera} 
translate images to different camera styles with CycleGAN \cite{zhu2017unpaired}, and then use both real and generated images to reduce inter-camera variation.
Ma \textit{et al.} 
%\cite{ma2017pose,ma2018disentangled}
\cite{ma2018disentangled}
 use a cGAN 
%\cite{mirza2014conditional}
 to generate pedestrian images with different poses to learn features free of influences of pose variation.
%
% Zheng \textit{et al} \cite{zheng2019joint} propose joint learning framework that end-to-end couples re-id learning and image generation in a unified network.
%
All those methods focus on single-modality RGB Re-ID and cannot deal with cross-modality RGB-IR Re-ID. Different from them, ours generate cross-modality paired-images and learn both set-level and instance-level aligned features.

\textbf{Image Translation.}
Generative Adversarial Network (GAN) \cite{goodfellow2014generative} learns data distribution in a self-supervised way via the adversarial training, which has been widely used in image translation. %\cite{isola2017image,zhu2017unpaired,choi2018stargan} and domain adaptation \cite{ganin2016domain,hoffman2018cycada}.
Pix2Pix \cite{isola2017image} solves the image translation by utilizing a conditional generative adversarial network and a reconstruction loss supervised by paired data.
CycleGAN \cite{zhu2017unpaired} and StarGAN \cite{choi2018stargan} learn images translations with unpaired data using cycle-consistency loss.
%
%In \cite{zhu2017unpaired}, with unpaired data, CycleGAN simultaneously learns two reciprocal image translations between two domains and enforces the translated images to reconstruct their original images.
%
%Further, StarGAN \cite{choi2018stargan} learns multi-domain image translations by making the generator take both images and domain labels as inputs, and improving the discriminator to simultaneously distinguish image sources and classify their domains.
%
Those methods only learn one-to-one mapping among different modalities and cannot be used in RGB-IR Re-ID, where the mapping from IR to RGB is one-to-many.
Different from them, our method first disentangles images to modality-invariant and modality-specific features, and then generates cross-modality paired-images by decoding from exchanged features.

%\textbf{Disentangled Feature Learning.}

\section{The Proposed Method}

Our method includes a generation module $\mathcal{G}$ to generate cross-modality paired-images and a feature alignment module $\mathcal{F}$ to learn both global set-level and fine-grained instance-level aligned features.
Finally, by training the two modules with re-id loss, we can learn both modality-aligned and identity-discriminative features.

\subsection{Cross-Modality Paired-Images Generation Module}

As shown in Figure \ref{fig:images_comparision}(b), in RGB-IR task, the training images from two modalities are unpaired, which makes it more difficult to reduce the gap between the RGB and IR modalities.
To solve the problem, we propose to generate paired-images by disentangling features and decoding from exchanged features. We suppose that images can be decomposed to modality-invariant and modality-specific features. Here, the former includes content information such as pose, gender, clothing category and carrying, \textit{etc}. Oppositely, the latter has style information such as clothing/shoes colors, texture, \textit{etc}. 
Thus, given unpaired-images, by disentangling and exchanging their style information, we can generate paired-images, where the two images have the same content information such as pose and view but with different style information such as clothing colors.

\textbf{Features Disentanglement.}
We disentangle features with three encoders.
The three encoders are the modality-invariant encoder $E^{i}$ of learning content information from both modalities, the RGB modality-specific encoder $E^{s}_{rgb}$ of learning RGB style information, and the IR modality-specific encoder $E^{s}_{ir}$ of learning IR style information.
Given RGB images $X_{rgb}$ and IR images $X_{ir}$, their modality-specific features $M^{s}_{rgb}$ and $M^{s}_{ir}$ can be learned in Eq.(\ref{eq:modality_specific_features}).
Similarly, their modality-invariant features $M^{i}_{rgb}$ and $M^{i}_{ir}$ can be learned in Eq.(\ref{eq:modality_invariant_features}).
\begin{equation}
M^{s}_{rgb} = E^{s}_{rgb}(X_{rgb}), \ M^{s}_{ir} = E^{s}_{ir}(X_{ir})
\label{eq:modality_invariant_features}
\end{equation}
\begin{equation}
M^{i}_{rgb} = E^{i}(X_{rgb}), \ M^{i}_{ir} = E^{i}(X_{ir})
\label{eq:modality_specific_features}
\end{equation}

\noindent\textbf{Paired-Images Generation.}
We generate paired-images using two decoders including a RGB decoder $D_{rgb}$ of generating RGB images and an IR decoder $D_{ir}$  of generating IR images.
After getting the disentangled features in Eq.(\ref{eq:modality_invariant_features}) and Eq.(\ref{eq:modality_specific_features}), we can generate paired-images by exchanging their style information.
Specifically, to generate RGB images $X_{ir2rgb}$ paired with real IR images $X_{ir}$, we can use the content features $M_{ir}^{i}$ from the real IR images $X_{ir}$ and the style features $M_{rgb}^{s}$ from the real RGB images $X_{rgb}$.
By doing so, the generated images will contain content information from the IR images and style information from the RGB image. Similarly, we can also generate fake IR images $X_{rgb2ir}$ paired with real RGB images $X_{rgb}$.
Note that to ensure that the generated images have the same identities with their original ones, we only exchange features  intra-person.
This processes can be formulated in Eq.(\ref{eq:decode}).
\begin{equation}
\begin{split}
X_{ir2rgb} = D_{ir}(M_{ir}^{i}, M_{rgb}^{s})
, X_{rgb2ir} = D_{rgb}(M_{rgb}^{i}, M_{ir}^{s})
\end{split}
\label{eq:decode}
\end{equation}

\noindent\textbf{Reconstruction Loss.}
A simple supervision is to force the disentangled features to reconstruct their original images. Thus, we can formulate the reconstruction loss $\mathcal{L}_{recon}$ as below, where $||\cdot||_{1}$ is L1 distance.
\begin{equation}
\begin{split}
\mathcal{L}_{recon} & = ||X_{rgb} - D_{rgb}(E^{i}(X_{rgb}), E^{s}_{rgb}(X_{rgb}))||_{1} \\
& + ||X_{ir} - D_{ir}(E^{i}(X_{ir}), E^{s}_{ir}(X_{ir})) ||_{1}
\end{split}
\label{loss:reconst}
\end{equation}

\noindent\textbf{Cycle-Consistency Loss.}
The reconstruction loss $\mathcal{L}_{recon}$ in Eq.(\ref{loss:reconst}) cannot supervise the cross-modality paired-images generation, and the generated images may not contain the expired content and style information. 
For example, when translating IR images $X_{ir}$ to its RGB version $X_{ir2rgb}$ via Eq(\ref{eq:decode}), the translated images $X_{ir2rgb}$ may not keep the poses (content information) from $X_{ir}$, or don't have the right clothing color (style information) with $X_{rgb}$. This is not the case we want and will harm the feature learning module.
Inspired by CycleGAN \cite{zhu2017unpaired}, we introduce a cycle-consistency loss to guarantee that the generated images can be translated back to their original version. By doing so, the consistency loss further limits the space of the generated samples.
The cycle-consistency loss can be formulated as below:
\begin{equation}
\mathcal{L}_{cyc} = || X_{rgb} - X_{rgb2ir2rgb} ||_{1} + || X_{ir} - X_{ir2rgb2ir} ||_{1}
\label{loss:cyc}
\end{equation}
where $X_{ir2rgb2ir}$ and $X_{rgb2ir2rgb}$ are the cycle-reconstructed images as in Eq.(\ref{eq:cycle_reconst_images}).
\begin{equation}
\begin{split}
X_{ir2rgb2ir} & = D_{ir}(E_{rgb}^{i}(X_{ir2rgb}), E_{ir}^{s}(X_{rgb2ir})) \\
X_{rgb2ir2rgb} & = D_{rgb}(E_{ir}^{i}(X_{rgb2ir}), E_{rgb}^{s}(X_{ir2rgb}) )
\end{split}
\label{eq:cycle_reconst_images}
\end{equation}

\noindent\textbf{GAN loss.}
The reconstruction loss $\mathcal{L}_{recon}$ and cycle-consistency loss $\mathcal{L}_{cyc}$ lead to blurry images. To make the generated images more realistic, we apply the adversarial loss \cite{goodfellow2014generative} on both modalities, which have been proved to be effective in image generation tasks \cite{isola2017image}.
Specifically, we import two discriminators $Dis_{rgb}$ and $Dis_{ir}$ to distinguish real images from the generated ones on RGB and IR modalities, respectively. In contrast, the encoders and decoders aim to make the generated images indistinguishable. The GAN loss can be formulated as below:
\begin{equation}
\begin{split}
\mathcal{L}_{gan} = & E_{}[log Dis_{rgb}(X_{rgb}) + log (1 - Dis_{rgb}({X}_{ir2rgb}))] \\
+ & E_{}[log Dis_{ir}(X_{ir})
 + log (1 - Dis_{ir}(X_{rgb2ir}))]
\end{split}
\end{equation}

\subsection{Feature Alignment Module}

\noindent\textbf{Set-Level Feature Alignment.}
To reduce the modality-gap, most methods attempt to learn a shared feature-space for different modalities by using dual path \cite{ye2018hierarchical,ye2018visible}, or GAN loss \cite{dai2018cross}.
However, those methods do not explicitly remove the modality-specific information, which may be encoded into the shared feature-space and harms the performance \cite{chang2019all}.
In our method, we utilize a set-level encoder $E^{sl}$ to learn set-level aligned features. The weights  $E^{sl}$ are shared with the modality-invariant encoder $E^{i}$. As we can see, in the cross-modality paired-images generation module, our modality-invariant encoder $E^{i}$ is trained to explicitly remove modality-specific features.
Thus, given images $X$ from any modality, we can learn their set-level aligned features $M=E^{sl}(X)$.

\noindent\textbf{Instance-Level Feature Alignment.}
Even so, as we discuss in the introduction, only performing global set-level alignment between the entire RGB and IR sets may lead to misalignment of some instances. To overcome this problem, we propose to perform instance-level alignment by using the cross-modality paired-images generated by the generation module.
Specifically, we first utilize instance-level encoder $E^{il}$ to map the set-level aligned features $M$ to a new feature space $\mathcal{T}$, \textit{i.e.} $T=E^{il}(M)$. Then, based on the feature space $\mathcal{T}$, we align every two cross-modality paired-images by minimizing their Kullback-Leibler Divergence. Thus, the loss of the instance-level feature alignment can be formulated in Eq.(\ref{eq:instance_level_align}).
\begin{equation}
\begin{split}
\mathcal{L}_{align} & = E_{(x_1, x_2) \in (X_{ir}, X_{ir2rgb})} [ KL(p_{1}||p_{2}) ]\\
& + E_{(x_1, x_2) \in (X_{rgb2ir}, X_{rgb})} [ KL(p_{1}||p_{2}) ]
\end{split}
\label{eq:instance_level_align}
\end{equation}
where $p_1=C(t_1)$ and $p_2=C(t_2)$ are the predicted probabilities of $x_1$ and $x_2$ on all identities, $t_1$ and $t_2$ are the features of $x_1$ and $x_2$ in the feature space $\mathcal{T}$, $C$ is a classifier implemented with a fully-connected layer.

\noindent\textbf{Identity-Discriminative Feature Learning.}
To overcome the intra-modality variation, following \cite{zheng2016person,hermans2017defense}, we averagely pool the feature maps $T$ in instance-level aligned space $\mathcal{T}$ to corresponding feature vectors $V$.
Given real images $X$, we optimize their feature vectors $V$ with a classification loss $\mathcal{L}_{cls}$ of a classifier $C$ and a triplet loss $\mathcal{L}_{triplet}$.
\begin{equation}
\mathcal{L}_{cls}  = E_{v \in V} (-log \ p(v))
\end{equation}
\begin{equation}
\mathcal{L}_{triplet} = E_{v \in V}[m - D_{v_a, v_p} + D_{v_a, v_n}]_{+}
\label{eq:triplet_loss}
\end{equation}
where $p(\cdot)$ is the predicted probability predicted by the classifier $C$ that the input feature vector belongs to the ground-truth,
$v_a$ and $v_p$ are a positive pair of feature vectors belonging to the same person, $v_a$ and $v_n$ are a negative pair of feature vectors belonging to different persons, $m$ is a margin parameter and $[x]_{+}=max(0, x)$.

\subsection{Overall Objective Function and Test}

Thus, the overall objective function of our method can formulated as below:
\begin{equation}
\begin{split}
\mathcal{L} & = \lambda_{cyc}\mathcal{L}_{cyc} + \lambda_{gan} \mathcal{L}_{gan} \\
& + \lambda_{align} \mathcal{L}_{align} + \lambda_{reid} (\mathcal{L}_{cls} + \mathcal{L}_{triplet})
\end{split}
\end{equation}
where $\lambda_{*}$ are weights of corresponding terms. Following \cite{zhu2017unpaired}, we set $\lambda_{cyc}=10$ and $\lambda_{gan}=1$. $\lambda_{reid}$ is set 1 empirically and $\lambda_{align}$ is decided by grid search.

During the test stage, only feature learning module $\mathcal{F}$ is used. Given images $X$, we use the set-level alignment encoder $E^{sl}$ and the instance-level encoder $E^{il}$ to extract features, \textit{i.e.} $V=E^{il}((E^{sl}(X))$. Finally, matching is conducted by computing cosine similarities of feature vectors $V$ between the probe images and gallery ones.

\section{Experiment}

\begin{table*}[!t]
\small
\center
\caption{Comparison with the state-of-the-arts on SYSU-MM01 dataset. The R1, R10, R20 denote Rank-1, Rank-10 and Rank-20 accuracies (\%), respectively. The mAP denotes mean average precision score (\%).
}
\setlength{\tabcolsep}{5.0pt}
\begin{tabular}{c|cccc|cccc|cccc|cccc}
\hline\hline
\multicolumn{1}{c}{\multirow{3}{*}{Methods}} & \multicolumn{8}{|c}{\textit{All-Search}} & \multicolumn{8}{|c}{\textit{Indoor-Search}} \\
\cline{2-17}
 &  \multicolumn{4}{|c}{\textit{Single-Shot}} &  \multicolumn{4}{|c|}{\textit{Multi-Shot}} &  \multicolumn{4}{c|}{\textit{Single-Shot}} &  \multicolumn{4}{c}{\textit{Multi-Shot}} \\
~ & R1 & R10 & R20 & mAP  & R1 & R10 & R20 & mAP  & R1 & R10 & R20 & mAP  & R1 & R10 & R20 & mAP \\
\hline
HOG & 2.76 & 18.3 & 32.0 & 4.24 &   3.82 & 22.8 & 37.7 & 2.16 &  3.22 & 24.7 & 44.6 & 7.25 &   4.75 & 29.1 & 49.4 & 3.51 \\
LOMO & 3.64 & 23.2 & 37.3 & 4.53   & 4.70 & 28.3 & 43.1 & 2.28   & 5.75 & 34.4 & 54.9 & 10.2 &   7.36 & 40.4 & 60.4 &  5.64 \\
Two-Stream &   11.7 & 48.0 & 65.5 & 12.9 &   16.4 & 58.4 & 74.5 & 8.03 &    15.6 & 61.2 & 81.1 & 21.5 &    22.5 & 72.3 & 88.7 & 14.0 \\
One-Stream &  12.1 &  49.7 & 66.8 & 13.7 &   16.3 & 58.2 & 75.1 & 8.59 &   17.0 & 63.6 & 82.1 & 23.0 &    22.7 & 71.8 & 87.9 & 15.1 \\
Zero-Padding &  14.8 & 52.2 & 71.4 & 16.0 &   19.2 & 61.4 & 78.5 & 10.9 &  20.6  & 68.4 & 85.8 & 27.0 &  24.5 & 75.9 & 91.4 & 18.7 \\
BCTR & 16.2 & 54.9 & 71.5 & 19.2 &     - & - & - & - &     - & - & - & - &     - & - & - & - \\
BDTR & 17.1 & 55.5 & 72.0 & 19.7 &     - & - & - & - &     - & - & - & - &     - & - & - & - \\
D-HSME & 20.7 & 62.8 & 78.0 & 23.2 &     - & - & - & - &     - & - & - & - &     -  & - & - & - \\
cmGAN & 27.0 & 67.5 & 80.6 & 27.8   & 31.5 & 72.7 & 85.0 & 22.3   & 31.7 & 77.2 & 89.2 & 42.2 &  37.0 & 80.9 & 92.3 & 32.8 \\
D$^2$RL & 28.9 & 70.6 & 82.4 & 29.2  & - & - & - & - &     - & - & - & - &     -  & - & - & - \\
\hline
{\textit{Ours}} & \textbf{38.1} & \textbf{80.7} & \textbf{89.9} & \textbf{36.9}   & \textbf{45.1} & \textbf{85.7} & \textbf{93.8} & \textbf{29.5}  & \textbf{43.8} & \textbf{86.2} & \textbf{94.2} & \textbf{52.9} &  \textbf{52.7} & \textbf{91.1} & \textbf{96.4} & \textbf{42.7}  \\
\hline
\hline
\end{tabular}
\label{table:state-of-the-art}
\end{table*}

\subsection{Dataset and Evaluation Protocol}

\noindent\textbf{Dataset}.
We evaluate our model on two standard
benchmarks including  SYSU-MM01 and RegDB.
(1) SYSU-MM01 \cite{wu2017rgb} is a popular RGB-IR Re-ID dataset, which includes 491 identities from 4 RGB cameras and 2 IR ones.  The training set contains 19,659 RGB images and 12,792 IR images of 395 persons and the test set contains 96 persons.
Following \cite{wu2017rgb}, there are two test modes, \textit{i.e.} \textit{all-search} mode and \textit{indoor-search} mode.
For the \textit{all-search} mode, all images are used. For the \textit{indoor-search} mode, only indoor images from $1st, 2nd, 3rd, 6th$ cameras are used. For both modes, the \textit{single-shot} and \textit{multi-shot} settings are adopted, where 1 or 10 images of a person are randomly selected to form the gallery set. Both modes use IR images as probe set and RGB images as gallery set.
(2) RegDB \cite{nguyen2017person} contains 412 persons, where each person has 10 images from a visible camera and 10 images from a thermal camera.

\noindent\textbf{Evaluation Protocols}. The Cumulative Matching Characteristic (CMC) and mean average precision (mAP) are used as evaluation metrics.
Following \cite{wu2017rgb}, the results of SYSU-MM01 are evaluated with official code based on the average of 10 times repeated random split of gallery and probe set.
Following \cite{ye2018hierarchical,ye2018visible}, the results of RegDB are based on the average of 10 times repeated random split of training and testing sets.

\subsection{Implementation Details}

In generation module $\mathcal{G}$, following \cite{radford2016unsupervised}, we construct our modality-specific encoders with 2 strided convolutional layers followed by a global average pooling layer and a fully connected layer. For decoders, following
%\cite{dumoulin2017a,ghiasi2017exploring,wang2017zm},
\cite{wang2017zm},
 we use 4 residual blocks with Adaptive Instance Normalization (AdaIN) and 2 upsampling with convolutional layers. Here, the parameters of AdaIN are dynamically generated by the modality-specific features. In GAN loss, we use discriminator and LSGAN as in \cite{mao2016least} to stable the training.

In feature learning module $\mathcal{F}$, for a fair comparison, we adopt the ResNet-50 \cite{he2016deep} pre-trained with ImageNet \cite{russakovsky2015imagenet} as our CNN backbone. Specifically, we use the first two layers of the ResNet-50 as our set-level encoder $E^{sl}$, and use the remaining layers as our instance-level encoder $E^{il}$.
For the classification loss, the classifier $C$ takes the feature vectors $V$ as inputs, followed by a batch normalization,
%\cite{ioffe2015batch}
a fully-connected layer and a soft-max layer to predict the inputs' labels.

\iffalse
We implement our model with open-source deep learning framework Pytorch.
%\footnote{https://pytorch.org/}.
The training images are resized to $256 \times 128$ and augmented with horizontal flip. The batch size is set to 128 (16 person, 4 RGB images and 4 IR images). We optimize our framework using Adam with learning rate 0.0002 and betas $[0.5, 0.999]$. The generation module is first pre-trained for 100 epochs. Then the overall framework is jointly optimized for 50 epochs, where the learning rate is decayed to its $0.1$ at 30 epochs. 
\fi

Please see our code \footnote{https://github.com/wangguanan/JSIA-ReID} for more details.

\subsection{Comparision with State-of-the-arts}

\textbf{Results on SYSU-MM01 Datasets.}
We compare our model with 10 methods including hand-crafted features (HOG \cite{dalal2005histograms}, LOMO \cite{liao2015person}), feature learning with the classification loss (One-Stream, Two-Stream, Zero-Padding) \cite{wu2017rgb}, feature learning with both classification and ranking losses (BCTR, BDTR) \cite{ye2018hierarchical}, metric learning (D-HSME \cite{hao2019hsme}), and reducing distribution divergence of features (cmGAN \cite{dai2018cross}, D$^2$RL \cite{wang2019learning}). The results are shown in Table \ref{table:state-of-the-art}.
Firstly, LOMO only achieves 3.64\%  and 4.53\% in terms of Rank-1 and mAP scores, respectively, which shows that hand-crafted features cannot be generalized to the RGB-IR Re-ID task.
Secondly, One-Stream, Two-Stream and Zero-Padding significantly outperform hand-crafted features by at least 8\%  and 8.3\% in terms of Rank-1 and mAP scores, respectively. This verifies that the classification loss contributes to learning identity-discriminative features.
Thirdly, BCTR and BDTR further improve Zero-Padding by 1.4\% in terms of Rank-1 and by 3.2\% in terms of mAP scores. This shows that the ranking and classification losses are complementary.
Additionally, D-HSME outperforms BDTR by 3.6\% Rank-1 and 3.5\% mAP scores, which demonstrates the effectiveness of metric learning.
In addition, D$^2$RL outperform D-HSME by 8.1\% Rank1 and 6.0\% mAP scores, implying the effectiveness of adversarial training.
Finally, Our method outperforms the state-of-the-art method by 9.2\% and 7.7\% in terms of Rank-1 and mAP scores, showing the effectiveness of our model.

% \subsection{Results on RegDB Dataset}

\begin{table}
\center
\caption{
Comparison with state-of-the-arts on the RegDB dataset under different query settings. }
\begin{tabular}{c|cc|cc}
\hline\hline
\multirow{2}{*}{Methods} & \multicolumn{2}{|c|}{thermal2visible} &  \multicolumn{2}{|c}{visible2thermal} \\
~ & Rank-1 & mAP & Rank-1 & mAP \\
\hline
Zero-Padding & 16.7 & 17.9 & 17.8 & 31.9 \\
TONE & 21.7 & 22.3 & 24.4 & 20.1 \\
BCTR &  - & - & 32.7 & 31.0 \\
BDTR & 32.8 & 31.2 & 33.5 & 31.9 \\
D$^2$RL & 43.4 & 44.1 & 43.4 & 44.1 \\
\hline
\textit{Ours} & 48.1 & 48.9 & 48.5 & 49.3 \\
\hline
\hline
\end{tabular}
\label{table:sota_regdb}
\end{table}

\textbf{Results on SYSU-RegDB Dataset.}
We evaluate our model on RegDB dataset and compare it with Zero-Padding \cite{wu2017rgb}, TONE \cite{ye2018visible}, BCTR \cite{ye2018hierarchical}, BDTR \cite{ye2018visible} and D$^2$RL \cite{wang2019learning}.
We adopt visible2thermal and thermal2visible modes. Here, the visible2thermal means that visible images are query set and thermal images are gallery set, and so on.
As shown in Table \ref{table:sota_regdb}, our model can significantly outperform the state-of-the-arts by 4.7\% and 5.1\% in terms of Rank-1 scores with thermal2visible and visible2thermal modes, respectively.
Overall, the results verify the effectiveness of our model.

\subsection{Model Analysis}

\begin{table}[t]
\center
\caption{Analysis of set-level (SL) and instance-level (IL) alignment. Please see text for more details.}
\begin{center}
\begin{tabular}{c|cc|cccc}
\hline\hline
index & SL & IL & R1 & R10 & R20 & mAP \\
\hline 
 1 & $\times$ & $\times$ &  32.1 & 75.7 & 87.0 & 31.9 \\
 2 & \checkmark & $\times$ & 35.1 & 78.6 & 88.2 & 33.8 \\
 3 & $\times$ & \checkmark & 36.0 & 79.8 & 89.0 & 35.5 \\
 4 & \checkmark & \checkmark & 38.1 & 80.7 & 89.9 & 36.9 \\
\hline
 5 & - & \checkmark & 36.8 & 80.2 & 89.4 & 36.0 \\
\hline
\hline
\end{tabular}
\end{center}
\label{table:ablation_study}
\end{table}

\textbf{Ablation Study.}
To further analyze effectiveness of the set-level alignment and the instance-level alignment, we evaluate our method under four different settings, \textit{i.e.} with or without set-level (SL) and instance-level (IL) alignment.
Specifically, when removing set-level alignment, we use separate set-level encoder $E^{sl}$, \textit{i.e.} we don't share weights of set-level encoder $E^{sl}$ with modality-invariant encoder $E^{i}$. When removing instance-level alignment, we set $\lambda_{align}=0$.
Moreover, to analyze whether the feature disentanglement contributes to set-level alignment, we remove the disentanglement strategy by using separate set-level encoder $E^{sl}$ and training it with a GAN loss as in \cite{dai2018cross}.

As shown in Table \ref{table:ablation_study}, when removing both SL and IL (index-1), our method only achieve $32.1\%$ Rank-1 score. 
By adding SL (index-2) or IL (index-3), the performance is improved to $35.1\%$ and $36.0\%$ Rank-1 score, which demonstrate the effectiveness of both SL and IL. 
When using both SL and IL (index-4), our method achieves the best performance at $38.1\%$ Rank-1 score, which demonstrates that SL and IL can be complementary with each other.  
Finally, when removing the disentanglement from set-level alignment (index-5), Rank-1 score drops by $1.3\%$. This illustrates that disentanglement is helpful for set-level alignment.

\begin{figure}[t]
\center
\includegraphics[scale=0.41]{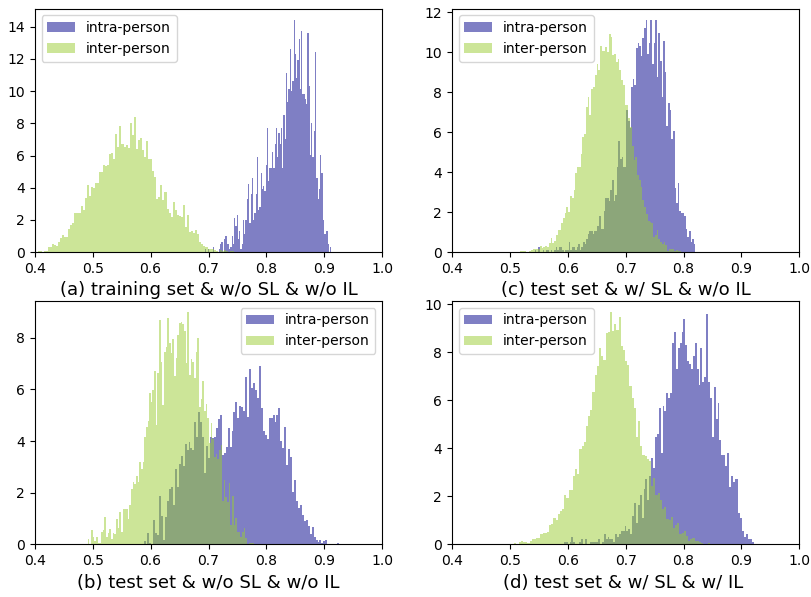}
\caption{Distribution of cross-modality similarities of intra-person and inter-person. The instance-level alignment (IL) can enhance intra-person similarity while keep inter-person similarity unchanged, which improves performance. w/ means with and w/o means without.}
\label{fig:distribution}
\end{figure}

To better understand set-level alignment (SL) and instance-level alignment (IL), we visualize the distribution of intra-person similarity and inter-person similarity under different variants. The similarity is calculated with cosine distance.
Firstly, when comparing with Figure \ref{fig:distribution}(a) and Figure \ref{fig:distribution}(b), we can find that even using no SL and IL, model can easily fit training set, while fails to generalize to test set. 
As we can see in Figure \ref{fig:distribution}(b), the two kind of similarities are seriously overlapped. This shows that the cross-modality variation cannot be well reduced by simply fitting identity information in training set.
Secondly, in Figure \ref{fig:distribution}(c), we find that although the similarity of intra-person becomes more concentrated, the similarity of inter-person also become larger. This shows that SL imports some misalignment of instances which may harm the performance.
Finally, in Figure \ref{fig:distribution}(c) we can see that, IL boosts intra-person similarity, meanwhile keeps the inter-person similarity unchanged.
In summary,  experimental results and analysis above show the importance and effectiveness of instance-level alignment.

\begin{figure}[t]
\center
\includegraphics[scale=0.36]{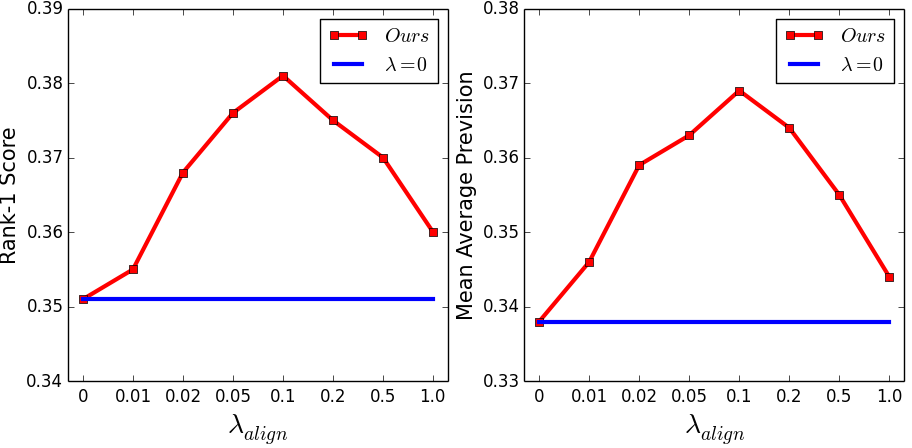}
\caption{Rank-1 and mAP scores with different $\lambda_{align}$ on SYSU-MM01 under \textit{single-shot\&all-search} mode.
}
\label{fig:parameters}
\end{figure}

\textbf{Parameters Analysis}.
We evaluate the effect of the weights, \textit{i.e.} $\lambda_{align}$. As shown in Figure \ref{fig:parameters}, we analyze our method with respect to the $\lambda_{align}$ on SYSU-MM01 dataset under \textit{single-shot\&all-search} mode. We can see that, with different $\lambda_{align}$, our method can stably have an significant improvement. The experimental results show that our method is robust to different weights.

\subsection{Visualization of Images}

\begin{figure}[t]
\center
\includegraphics[width=\linewidth]{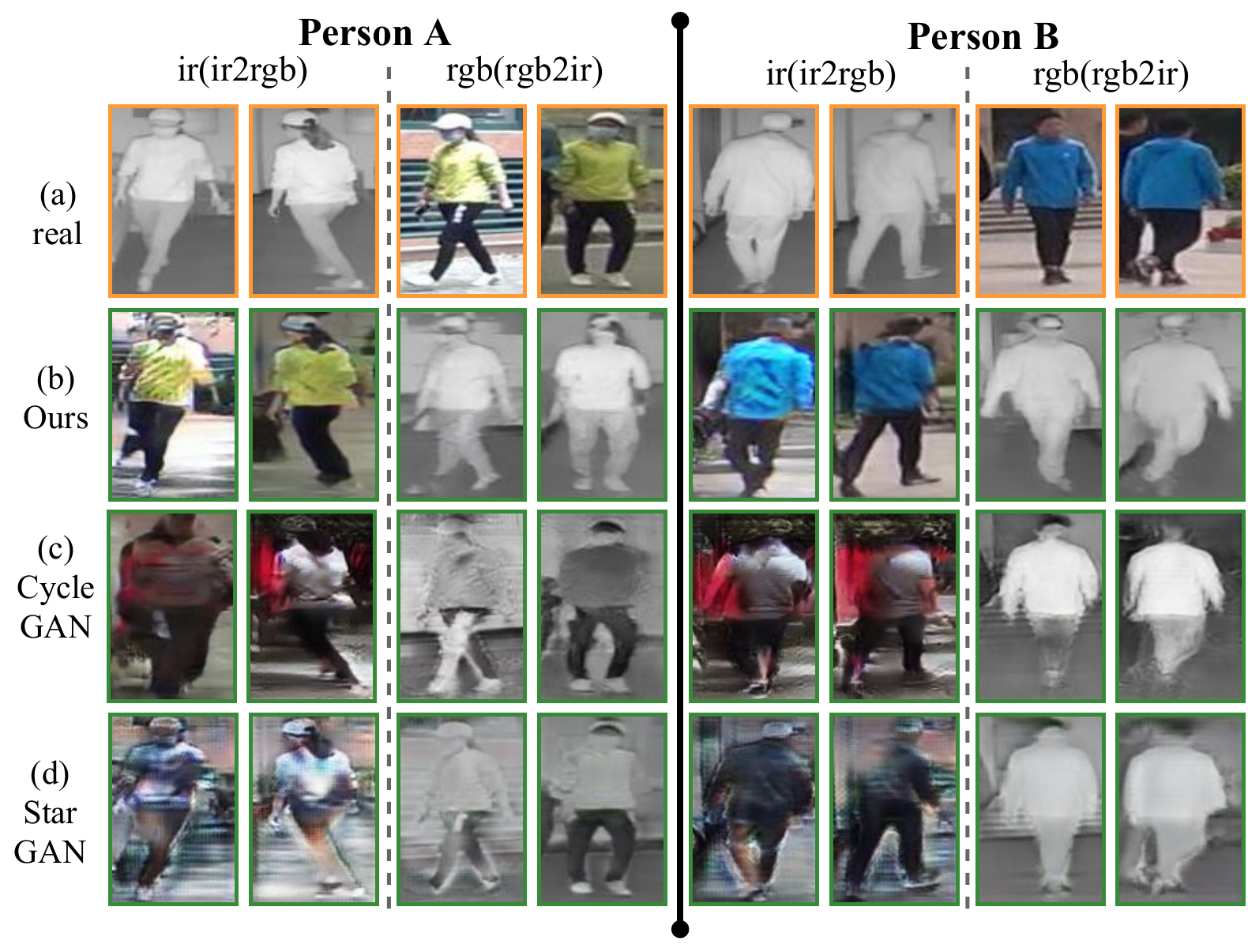}
\caption{Comparision among generated images from ours, CycleGAN \cite{zhu2017unpaired} and StarGAN \cite{choi2018stargan}. Ours can stably generate paired-images with given real ones, while CycleGAN and StarGAN fail.
}
\label{fig:fake_images}
\end{figure}

In this part, we display the generated cross-modality paired-images from ours, CycleGAN \cite{zhu2017unpaired} and StarGAN \cite{choi2018stargan}. From Figure \ref{fig:fake_images}(a), we can see that, images of a person in the two modalities are significant different, even human beings cannot easily identify them. In Figure \ref{fig:fake_images}(b), our method can stably generate fake images when given cross-modality unpaired-images from a person. For example, in person A, ours can translate her IR images to RGB version with right colors (yellow upper and black bottom clothes). However, in Figure \ref{fig:fake_images}(c) and Figure \ref{fig:fake_images}(d), CycleGAN and StarGAN cannot learn the right colors even poses. For example, person B should have blue upper clothing. However, images generated by CycleGAN and StarGAN are red and black, respectively. Those unsatisfying images cannot be used to learn instance-level aligned features.

\section{Conclusion}

In this paper, we propose a novel Joint Set-Level and Instance-Level Alignment Re-ID (JSIA-ReID).
On the one hand, our model performs set-level alignment by disentangling modality-specific and modality-invariant features. Compared with vanilla methods, ours can explicitly remove the modality-specific information and significantly reduce the modality-gap.
On the other hand, given cross-modality unpaired images, we we can generate cross-modality paired-images by exchanging their features. With the paired-images, instance-level variations can be reduced by minimizing the distances between every pair of images.
%
% Finally, together with re-id loss, our model can learn both modality-aligned and identity-discriminative features.
%
Experimental results on two datasets show the effectiveness of our proposed method.

\section{Acknowledgments}

This work was supported in part by the National Natural Science Foundation of China under Grants 61720106012, 61533016 and 61806203, the Strategic Priority Research Program of Chinese Academy of Science under Grant XDBS01000000, and the Beijing Natural Science Foundation under Grant L172050.

\bibliographystyle{aaai}
\bibliography{egbib}

\bigskip
\end{document}